\documentclass[a4paper,conference,a4paper]{IEEEtran}
\ifCLASSINFOpdf
\else
\fi

\usepackage{graphicx}
\usepackage{amsmath, amssymb, amsthm}
\usepackage{float}
\usepackage{adjustbox} 
\usepackage{xcolor}
\usepackage{multirow}
\usepackage{makecell}
\usepackage{textcomp}
\usepackage{subfig}

\DeclareMathOperator{\Tr}{tr}
\newcommand{\frmat}[2] {\mathbb{R}_*^{#1 \times #2}}
\def \calO {\mathcal{O}}
\newcommand{\psd}[2] {\mathcal{S}^+(#1,#2)}

\begin{document}
%
\title{Automatic Estimation of Self-Reported Pain by Interpretable Representations of Motion Dynamics}



%
\author{\IEEEauthorblockN{Benjamin Szczapa\IEEEauthorrefmark{1},
Mohamed Daoudi\IEEEauthorrefmark{2},
Stefano Berretti\IEEEauthorrefmark{3},
Pietro Pala\IEEEauthorrefmark{3}, Alberto Del Bimbo\IEEEauthorrefmark{3} and Zakia Hammal\IEEEauthorrefmark{4}}
\IEEEauthorblockA{\IEEEauthorrefmark{1}Univ. Lille, CNRS, Centrale Lille, UMR 9189 CRIStAL, F-59000 Lille, France}
\IEEEauthorblockA{\IEEEauthorrefmark{2}IMT Lille Douai, Univ. Lille, CNRS, UMR 9189 CRIStAL, F-59000 Lille, France}
\IEEEauthorblockA{\IEEEauthorrefmark{3}Department of Information Engineering, University of Florence, Italy}
\IEEEauthorblockA{\IEEEauthorrefmark{4}Robotics Institute, Carnegie Mellon University, Pittsburgh, PA, USA}}


\maketitle

\begin{abstract}
We propose an automatic method for pain intensity measurement from video. For each video, pain intensity was measured using the dynamics of facial movement using 66 facial points. Gram matrices formulation was used for facial points trajectory representations on the Riemannian manifold of symmetric positive semi-definite matrices of fixed rank. Curve fitting and temporal alignment were then used to smooth the extracted trajectories. A Support Vector Regression model was then trained to encode the extracted trajectories into ten pain intensity levels consistent with the Visual Analogue Scale for pain intensity measurement. The proposed approach was evaluated using the UNBC McMaster Shoulder Pain Archive and was compared to the state-of-the-art on the same data. Using both 5-fold cross-validation and leave-one-subject-out cross-validation, our results are competitive with respect to state-of-the-art methods.
\end{abstract}


%
\IEEEpeerreviewmaketitle

\section{Introduction}
Pain is an unpleasant sensory and emotional experience associated with actual or potential tissue damage and caused by illness or injury~\cite{Merskey:1979}. The assessment of pain is accomplished primarily through subjective self-report using the Visual Analog Scale (VAS) or the Numerical Rating Scale (NRS)~\cite{Younger2009PainOA}. The most commonly used scale in clinical assessment is the VAS~\cite{Aicher:2012, Farrar:2000, Jensen:2003, Jensen:2005}. However, while useful, self-reported pain is difficult to interpret and may be impaired or, in some circumstances, not possible to obtain (\emph{e.g.}, for children or patients requiring breathing assistance).

Significant efforts have been made in human behavioral studies to identify reliable and valid facial indicators of pain~\cite{Craig:2011, Kunz:2007, Prkachin:1992, Prkachin:2008}. In these studies, pain expression and intensity were reliably characterized at the frame level by the activation of a set of anatomical facial actions using the manual Facial Action Coding System (FACS)~\cite{Ekman:2002}. However, manual FACS based pain assessment requires over a hundred hours of training for FACS certification, and approximately an hour or more to manually annotate a minute of video. The intensive time required to annotate videos using the FACS makes it ill suited for real-time application and clinical use. A powerful alternative to manual annotation is the automatic and objective assessment of pain from facial expression~\cite{Hammaletal2018}.

The last decade has witnessed an increasing effort to address the need for an automatic, objective, and efficient measurement of pain from video. Most previous efforts in automatic assessment of pain have focused on pain detection or pain intensity estimation at the frame-level (see~\cite{Hammaletal2018} and~\cite{Werner:2019} for a detailed review of previous efforts on the topic).

\begin{figure*}[!ht]
\centering
\includegraphics[width=0.85\textwidth]{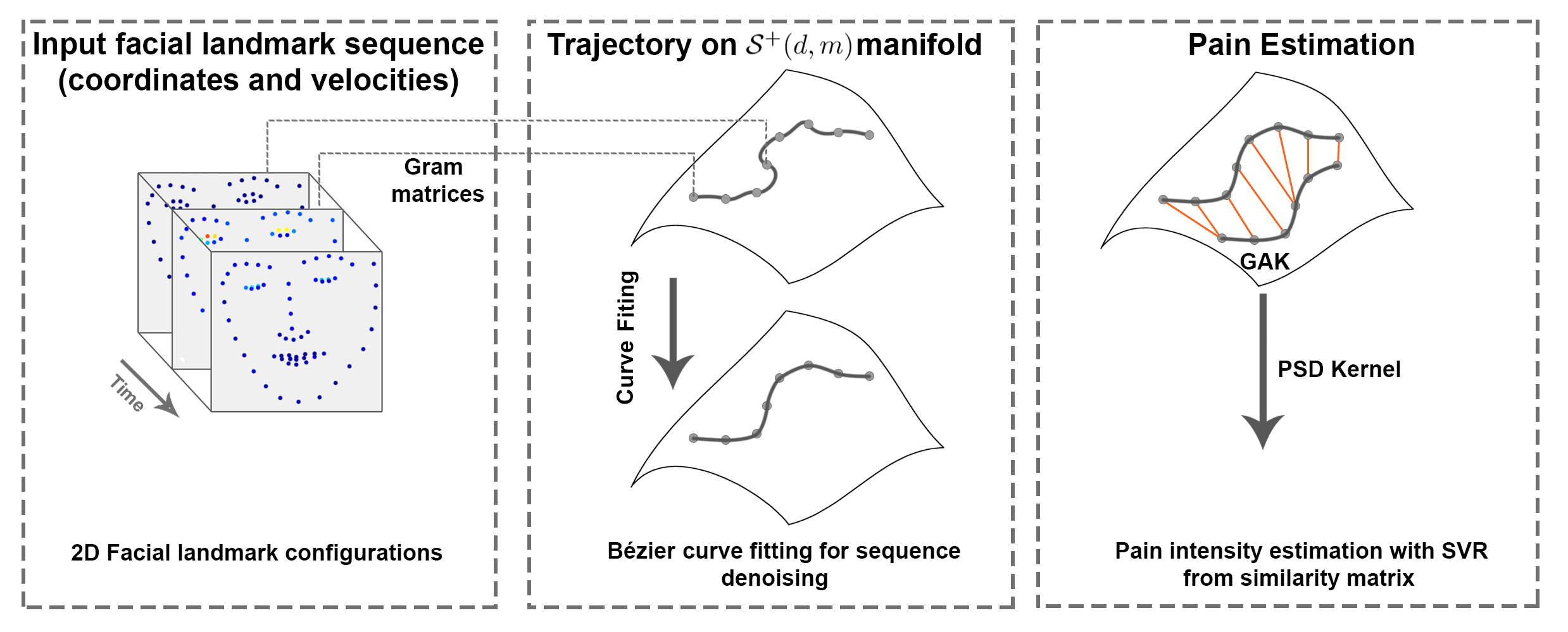}
\caption{Overview of the proposed approach: (left plot) First, facial landmarks are detected using Active Appearance Model (AAM) on each video frame and velocities are computed as the displacement of the coordinates between two consecutive frames. Then Gram matrices are computed from the combination of the landmark coordinates and velocities. These matrices delineate a trajectory on the $\mathcal{S}^+(d,m)$ manifold; (middle plot) We apply a curve fitting algorithm to the trajectory for smoothing and noise reduction;
(right plot) The Global Alignment Kernel (GAK) is then used to align the trajectories on the manifold, which results in a similarity score between the trajectories. Finally, we use the kernel generated from GAK with SVR to estimate the pain intensity.}
\label{fig:overview}
\end{figure*}

A few recent exceptions~\cite{Martinez:cvrpw17, LiuPRP17-DeepFaceLIFT}, have investigated video based pain intensity measurement consistent with self-reported VAS. The VAS is a self-reported pain scale that indicates pain experience on a 0 to 10 scale (where 0 is for "no pain" and 10 is for "worse possible pain"). For instance, using the UNBC-McMaster Shoulder Pain Archive database~\cite{UNBC-McMaster}, Martinez \emph{et al.}~\cite{Martinez:cvrpw17} proposed a two step learning approach to estimate pain consistent with the VAS. The authors employed a Recurrent Neural Network (RNN) to first estimate pain score at frame level. The estimated scores were then fed into a personalized Hidden Conditional Random Fields (HCRF) to estimate pain score at the video level consistent with the VAS. Using the same pain database, Liu \emph{et al.}~\cite{LiuPRP17-DeepFaceLIFT} proposed a two-stage personalized model, named DeepFaceLIFT, for automatic estimation of the self-reported VAS score. The authors used a Neural Network and Gaussian process regression model and combined facial expression and a set of hand-crafted personal features for pain score measurement at the video level.

Previous efforts for video based pain assessment used artificial neural networks to first estimate pain score at the frame level before combining them to estimate pain score at the video level. We propose to extend previous work in video based assessment of pain intensity by estimating VAS score directly from video using a geometry based approach. To capture changes in the dynamics of facial movement relevant to pain expression, we propose an original framework based on Gram matrix computation and trajectory modeling on the Riemannian manifold of symmetric positive-semidefinite (PSD) matrices~\cite{Szczapa2019ICCVW}. With this representation, pain estimation is modeled as a problem of computing similarity between trajectories on the manifold using Support Vector Regression~\cite{DruckerBKSV96SVR}. 

\section{Face Representation}\label{sec:gram-matrix}
We propose a video based measurement of pain intensity scores using the dynamics of facial movement. Figure~\ref{fig:overview} shows an overview of the proposed approach. Given a set of $n_{seq}$ sequences, we first build the trajectories on the manifold $\psd{d}{m}$ from the Gram matrices of each frame of each sequence using the landmark configurations (and their velocities) as input features. We then compute the distances between all the trajectories and build a kernel $K$ that contains all the similarity scores after aligning the trajectories with the Global Alignment Kernel (GAK). Finally, we estimate pain intensity score based on the similarity matrix.

\subsection{Facial Shape Representation}

Given an image sequence $Is$, we represent the dynamics of facial movements with a time series formed by the coordinates $(x,y)$ of $n$ tracked facial landmarks. At a generic time (frame) $f$, facial expression is represented by a configuration $Z \in \mathrm{R}^{n \times 2}$ composed of $n$ tracked facial landmarks $p_i=(x_i,y_i)$, where $i \in \{1, \ldots, n\}$.
Thus, an image sequence is represented by a sequence of configuration matrices $Is = \{Z_1,\ldots,Z_{f}, \ldots Z_{\tau}\}$ with $f$ denoting the frame number and $\tau$ the number of frames of the sequence $Is$.
In addition to landmark coordinates, we compute for each landmark $p_i$ its velocity as the magnitude of the displacement between two consecutive landmark configurations $Z_f$ and $Z_{f+1}$. We denote the velocity matrix at frame $F$ as $V_F = Z_{f+1} - Z_f \in \mathrm{R}^{n \times 2}$, with $F \in \{1, \ldots, \tau-1\}$.
The final facial representation $R$ is the concatenation of the configuration matrix $Z$ and the velocity matrix $V$, where $R = [Z;V] \in \mathrm{R}^{2n \times 2}$.

We aim to measure the dynamic changes of the curves made of landmark configurations, while remaining invariant to rigid transformations like rotations and translations. Invariance to rigid transformation within each frame is obtained by computing coordinates of landmarks (and their velocities) as offsets with respect to the center of the face that is measured as the arithmetic mean of the landmarks: 
\begin{equation}
\label{eq:normalization}
(\bar{x_i}, \bar{y_i}) = \frac{1}{n}\sum_{i=1}^{n} (x_i, y_i) \;.
\end{equation}

\noindent 

We denote $A$ the normalized facial configuration of matrix $R$. Similarly to~\cite{Szczapa2019ICCVW,KacemDABP20}, this representation is further refined by extracting the Gram matrix $G$, which is the inner product of each facial configuration matrix as:
\begin{equation}
\label{eq:gram}
G = AA^{T} =\left\langle p_{i}, p_{j}\right\rangle, \quad 1 \leq i, j \leq 2n \;.
\end{equation}

In the following, we denote $m=2n$ the size of the facial configuration matrix for simplicity.

\subsection{Riemannian Geometry of Gram Matrix}
\label{sec:metric}
Given that each Gram matrix represents the landmarks configuration at the frame level, we propose (1) a geometry of space to model the dynamic changes of landmarks during a video sequence, and (2) a metric that allows to compute the distance between consecutive Gram matrices. In the following, we present a general metric that works for both 2D or 3D data and an optimized metric for 2D data.

Gram matrices are $m \times m$ positive-semidefinite (PSD) matrices of rank smaller than or equal to $d$ (in our case the rank is always equal to $d$). In this representation, $d$ is the dimensionality of the space where each landmark lies (i.e.,  $d=2$ for 2D landmarks and $d=3$ for 3D landmarks). We consider here the Riemannian geometry of the space $\psd{d}{m}$ of $m \times m$ positive-semidefinite matrices of rank $d$. This Riemannian geometry has been studied in~\cite{bonnabel:2009, journee2010low, massart2020quotient, massart2019curvature, vandereycken2009embedded, vandereycken2013riemannian} and used in~\cite{faraki:2016, meyer2011regression, gousenbourger2017piecewise, massart2019interpolation}.
In order to develop algorithms on the manifold, we resort to first order local approximations on the manifold. These approximations are called the \textit{tangent spaces}. This requires two fundamental tools: the Riemannian \textit{logarithm}, that maps points from the manifold to the tangent space, and the Riemannian \textit{exponential} that allows us to map tangent vectors from the tangent space to the manifold.

We consider here the manifold of $\psd{d}{m}$ as the quotient manifold $\frmat{m}{d} / \calO_d$, where $\frmat{m}{d}$ is the set of full-rank $m \times d$ matrices and $\calO_d$ is the orthogonal group in dimension $d$. 
The identification of $\psd{d}{m}$ with the quotient $\frmat{m}{d} / \calO_d$ comes from the following observation: Any PSD matrix $G \in \psd{d}{m}$ can be factorized as $G = AA^T$, with $A \in \frmat{m}{d}$. However, this factorization is not unique, as any matrix $\tilde A := AQ$, with $Q \in \calO_d$, satisfies $\tilde A \tilde A^T = A QQ^T A^T = G$. The two points $A$ and $\tilde A$ are thus \emph{equivalent} with respect to this factorization, and the set of equivalent points:
\begin{equation*}
A \calO_d := \{  AQ | Q \in \calO_d \} \; ,
\end{equation*}

\noindent
is called the equivalence class associated to $G$. The quotient manifold $\frmat{m}{d} / \calO_d$ is defined as the set of equivalence classes. The mapping $\pi : \frmat{m}{d} \to \frmat{m}{d} / \calO_d$, between points and their equivalence class, induces a Riemannian metric on the quotient manifold from the Euclidean metric in $\frmat{m}{d}$. This metric results in the following distance between PSD matrices ~\cite{massart2020quotient} :
\begin{equation}
\label{eq:PSD-norm}
d(G_i, G_j) = \mathrm{tr}(G_i) + \mathrm{tr}(G_j) - 2 \mathrm{tr}\left( \left( G_i^{\frac{1}{2}}  G_j G_i^{\frac{1}{2}} \right)^{\frac{1}{2}} \right) \; .
\end{equation}

\noindent
This distance can be expressed in terms of the facial configurations $A_i, A_j \in \frmat{m}{d}$ as follows:
\begin{equation}
\label{eq:frob-norm-new-metric}
d(G_i, G_j) = \min_{Q \in \calO_d} \lVert A_jQ-A_i\rVert_F \; ,
\end{equation}

\noindent
where $\lVert .\rVert_F$  is the Frobenius norm. The optimal solution is $Q^* := VU^{T}$, where $A_i^{T} A_j = U \Sigma V^{T}$ is a singular value decomposition.

In the specific case of 2D landmarks, when $d = 2$, the distance can be reformulated. Considering $G_i, G_j \in \psd{2}{m}$ to be two Gram matrices obtained from facial configurations $A_i, A_j \in \mathbb{R}^{m \times 2}$, the Riemannian distance~\eqref{eq:PSD-norm} can be expressed as:
\begin{equation}
\label{eq:distp2}
d(G_i, G_j)= \Tr(G_i) + \Tr(G_j) - 2\sqrt{(a + d)^2 + (c - b)^2} \; ,
\end{equation}

\noindent
where $A_i^TA_j$ = $\left( \begin{array}{cc} a & b \\ c & d \end{array} \right)$. The interested readers can find the proof of this expression in~\cite[\S 9]{Szczapa2019ICCVW}.

\section{Representation of Face Dynamics}\label{sec:traj-modeling}

\subsection{Trajectory Modeling}
The dynamic changes of facial landmarks movement originate trajectories on the Riemannian manifold of positive-semidefinite matrices of fixed rank. More specifically, we fit a curve $\beta_G$ to a sequence of facial configurations $\{A_0, \ldots, A_\tau\}$ represented by their corresponding Gram matrices $\{G_0,\ldots,G_\tau\}$ in $\psd{d}{m}$. This curve enables us to model the spatio-temporal evolution of the elements on $\psd{d}{m}$. Modeling a sequence of landmarks as a piecewise-geodesic curve on $\psd{d}{m}$ showed very promising results when the data are well acquired, \emph{i.e.}, without tracking errors or missing data. To account for both missing data and tracking errors, we rely on a more recent curve fitting algorithm: fitting by composite cubic blended curves~\cite[\S 5]{Gousenbourger2018}. Specifically, given a set of points $\{G_0,\ldots,G_\tau\} \in \psd{d}{m}$ associated to times $\{t_0, \dots, t_\tau\}$, with $t_i := i$, the curve $\beta_G$, defined on the interval $[0,\tau]$, is defined as:
\begin{equation}
\beta_G(t) := \gamma_i(t-i), \qquad t \in [i, i+1] \; ,
\end{equation}

\noindent
where each curve $\gamma_i$ is obtained by blending together fitting cubic B\'ezier curves computed on the tangent spaces of the data points $d_i$ and $d_{i+1}$ (represented by Gram matrices on the manifold).

These fitting cubic B\'ezier curves depend on a parameter $\lambda$, allowing us to balance two objectives: (1) proximity to the data points at the associated time instants, and (2) regularity of the curve (measured in terms of mean square acceleration). A high value of $\lambda$ results in a curve with possibly high acceleration that almost interpolates the data, while taking $\lambda \rightarrow 0$ results in a smooth function approximating the original trajectory.

\subsection{Global Alignment}
As explained in the previous section, we represent a sequence as a trajectory of Gram matrices in $\psd{d}{mn}$. Because videos could be of different duration (i.e., in our case video sequences of pain), the length of corresponding trajectories represented in this manifold can be different. 
A commonly used method to compute the similarity between trajectories with different length is Dynamic Time Warping (DTW). However, DTW does not define a proper metric and cannot be used to derive a valid positive-definite kernel. This would hamper the use of many approaches (including Support Vector Regression) to learn the mapping between trajectories in $\psd{d}{m}$ and pain intensity.
Cuturi \emph{et al.}~\cite{CuturiVBM07} proposed the Global Alignment Kernel (GAK) to address non-positive definite kernel defined by DTW. GAK allows to derive a valid positive-definite kernel when aligning two time series. As opposed to the DTW, the GAK generated kernel, that is the similarity matrix between all the sequences, can be used directly with Support Vector Regression. In fact, the kernels built with DTW do not show favorable positive definiteness properties as they rely on the computation of an optimum rather than the construction of a feature map. In terms of complexity, similar to naive implementation of DTW, the computational complexity of the GAK kernels is quadratic.

Let us now consider $G^1= \{G^1_0, \cdots, G^1_{\tau_1}\}$ and $G^2=\{G^2_0, \cdots, G^2_{\tau_2}\}$, two trajectories of Gram matrices. Given a metric to compute the distance between two elements of each sequence, we propose to compute the matrix $D$ of size $\tau_1 \times \tau_2$, where each $D(i,j)$ is the distance between two elements of the sequences, with $1 \leq i \leq \tau_1$ and $1 \leq j \leq \tau_2$:
\begin{equation}
\label{eq:matrix-distance}
D(i,j) = d(G^1_i, G^2_j) \; .
\end{equation}

The kernel $\tilde{k}$ can now be computed using the halved Gaussian Kernel on this same matrix $D$. Therefore, the kernel $\tilde{k}$ can be defined as:
\begin{equation}
\label{eq:k-tilde}
\tilde{k}(i,j) = \frac{1}{2} * exp\left(-\frac{D(i,j)}{\sigma^2}\right) \; .
\end{equation}

\noindent
As reported in~\cite{CuturiVBM07}, we can redefine our kernel as:
\begin{equation}
\label{eq:kernel-PSD}
k(i,j) = \frac{\tilde{k}(i,j)}{(1-\tilde{k}(i,j))} \; .
\end{equation}

\noindent
This strategy guarantees that the kernel is positive-semidefinite and can be used in its own. Finally, we can compute the similarity score between the two trajectories $G^1$ and $G^2$. This computation is performed in quadratic complexity, like DTW. To do so, we define a new matrix $M$ that contains the path to the similarity between our two sequences. We define $M$ as a zeros matrix of size $(\tau_1+1) \times (\tau_2+1)$ and $M_{0,0} = 1$. Computing the terms of $M$ is done using Theorem~2 in~\cite[\S 2.3]{CuturiVBM07}:
\begin{equation}
\label{eq:matrix-similarity}
M_{i,j} = (M_{i,j-1} + M_{i-1,j-1} + M_{i-1,j})*k(i,j) \; .
\end{equation}

\noindent
The similarity score between the trajectories $G^1$ and $G^2$ is given by the value at $M_{(\tau_1+1),(\tau_2+1)}$.

\section{Pain Estimation with Support Vector Regression}\label{sec:pain}
We build a new matrix $K$ of size $n_{seq} \times n_{seq}$, where $n_{seq}$ is the number of sequences in the dataset used to test our method. This symmetric matrix contains all the similarity scores between all the sequences of the dataset. This matrix is built with values computed from positive-semidefinite kernel, meaning that it is a positive-semidefinite matrix itself. Now that we have a valid and positive-semidefinite kernel $K$, as demonstrated by Cuturi \emph{et al.}~\cite{CuturiVBM07}, we can use it directly as a valid kernel for classification. To estimate pain intensity score (i.e., self-reported VAS scores), we use a Support Vector Regression (SVR) model. To train our SVR model, we give as input a training set that is a part of our kernel $K$ containing the similarity scores between all training trajectories. This part of the kernel, containing the training set, is also positive-semidefinite by definition. We also give a vector containing the labels for the trajectories in our training kernel. 
Because pain scores are continuous, to test the performance of our method, we compute the Mean Absolute Error (MAE) between the estimated pain scores and the ground truth (i.e., self-reported VAS pain scores). The MAE is computed as follows:
\begin{equation}
\mathrm{MAE}=\frac{1}{n_{seq}} \sum_{i=1}^{n_{seq}}\left|y_{i}-x_{i}\right| \; ,
\end{equation}

\noindent
where $n_{seq}$ is the number of sequences in the dataset, $y_{i}$ is the ground truth (i.e., self-reported VAS pain score), and $x_{i}$ is the predicted pain score.

\section{Experimental Results}\label{sec:results}
The UNBC-McMaster Shoulder Pain Archive ~\cite{UNBC-McMaster} was used to evaluate the reliability of the proposed approach for pain intensity measurement from the dynamics of facial landmark sequences. We used MatLab for the code and the Manopt library~\cite{manopt}.

\subsection{The UNBC-McMaster Shoulder Pain Archive}\label{sec:dataset}

\begin{figure}[!ht]
\centering
\subfloat[]{\label{fig:frame-34}\includegraphics[width=0.4\linewidth]{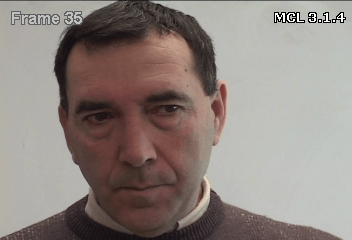}}
~
\subfloat[]{\label{fig:frame-34-landmarks}\includegraphics[width=0.4\linewidth]{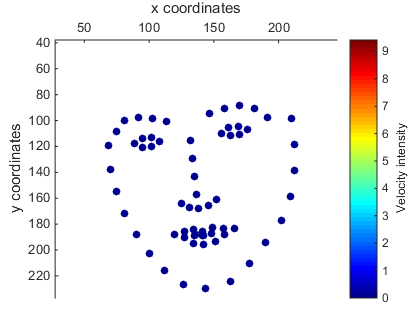}}
~ \\
\subfloat[]{\label{fig:frame-289}\includegraphics[width=0.4\linewidth]{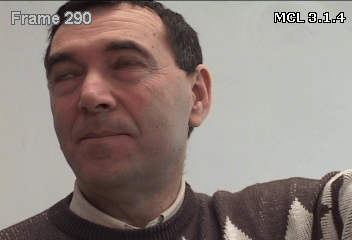}}
~
\subfloat[]{\label{fig:frame-289-landmarks}\includegraphics[width=0.4\linewidth]{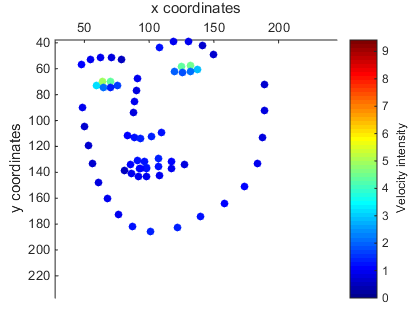}}
\caption{Example images from the UNBC-McMaster Shoulder Pain Archive in (a) and (c). In (b) and (d) their corresponding landmark coordinates and velocities, respectively (best viewed in color)~\cite{UNBC-McMaster}. }
\label{fig:example-frames-dataset}
\end{figure}
The UNBC-McMaster Shoulder Pain Archive dataset~\cite{UNBC-McMaster} is a widely used dataset for pain expression recognition and intensity estimation. The dataset contains 200 facial videos of 25 different subjects performing a series of active and passive range-of-motion of their affected and unaffected shoulders. Each video sequence is annotated for pain intensity score using three self-reported scales (including the VAS) and an Observer Pain Rating scale. The sequences are also annotated at the frame-level using the manual FACS (Facial Action Coding System). Figure~\ref{fig:example-frames-dataset} shows two images from a sequence of the dataset with their corresponding facial landmark representations and velocities. Our goal is to estimate pain intensity scores consistent with the VAS.
Table~\ref{tab:vas-distribution} shows the distribution of the VAS scores across the dataset. We can observe that the number of sequences are not the same for all the VAS scores.

\begin{table}[!ht]
\centering
\caption{Distribution of the VAS pain scores in the UNBC-McMaster Shoulder Pain Archive}
    \begin{tabular}{c|c}
        \textbf{VAS Score} & \textbf{Number of Sequences} \\ \hline \hline
        0 & 35 \\ \hline
        1 & 42 \\ \hline
        2 & 24 \\ \hline
        3 & 20 \\ \hline
        4 & 21 \\ \hline
        5 & 11 \\ \hline
        6 & 11 \\ \hline
        7 & 6 \\ \hline
        8 & 18 \\ \hline
        9 & 10 \\ \hline
        10 & 2 \\ \hline
    \end{tabular}
\label{tab:vas-distribution}
\end{table}

\begin{figure}[!ht]
\centering
\includegraphics[width=0.8\linewidth]{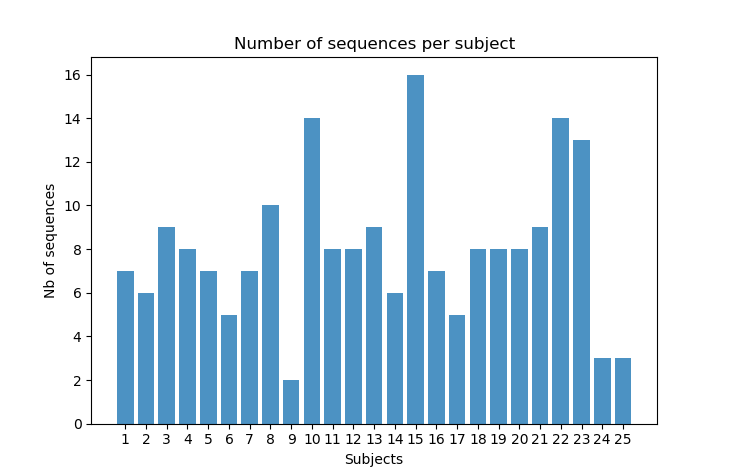}
\caption{Number of sequences per subjects in the UNBC-McMaster Shoulder pain archive dataset.}
\label{fig:sequence-distribution}
\end{figure}

Figure~\ref{fig:sequence-distribution} shows the number of sequences per subject. We can observe some disparity between the subjects that may represent a challenge for training as the number of sequences used will not be consistent across the dataset.

\subsection{Evaluation protocols}
We used three different protocols to evaluate the proposed method: Leave-One-Sequence-Out cross validation, Leave-One-Subject-Out cross validation, and 5-fold cross validation.

\paragraph{Leave-One-Sequence-Out cross validation protocol} In this protocol, training and testing are done on different sequences. For each round, we use all sequences of the dataset but one for training and the remaining sequence for testing. That is, data from the same subject can be used during the training and the testing phase as there are at least two sequences per subject in the dataset. Therefore, this protocol is sequence-independent but not subject-independent. We use this protocol as a baseline for our approach. 

\paragraph{Leave-One-Subject-Out cross validation protocol} In this second protocol, for each round, we use all the sequences from all subjects but one for training, and the remaining subject for testing (no overlap between the training subjects and test subject). We do this operation for all the subjects (i.e., 25 rounds) in the dataset, so that in the end each subject will be used for testing once. 

\paragraph{5-fold cross validation protocol} This third protocol is similar to the Leave-One-Subject-Out cross validation protocol, but instead of taking only the sequences of one subject at a time for testing, we take all the sequences of five subjects for testing and the remaining sequences for the training. To choose the five subjects for testing, we choose the five first subjects in the dataset, then the five next subjects and so on until all the subjects are used for testing.

The advantage of using cross validation is to prevent from having performance results that are due to the chance (all data will be used to train and test the proposed method). The average across all folds is more representative of the whole dataset. 

\subsection{Pain estimation from landmark coordinates and velocities}
Our goal is to estimate the VAS pain score for each sequence of the dataset. We test our method with the three protocols described above and report the results in Table~\ref{tab:results-velocities}. For each protocol, we fix the value of the curve fitting parameter lambda to 1000 and the Gaussian kernel in the sequence alignment sigma to 0.8 (see Table~\ref{tab:results-velocities}). \emph{Protocol} indicates the protocol used for training and testing our method; \emph{\% of frames} indicates the percentage of frames used from each sequence for training and testing; \emph{MAE} indicates the Mean Absolute Error and \emph{RMSE} the Root Mean Square Error of our estimation (see Table~\ref{tab:results-velocities}).

\begin{table}[!htb]
\centering
\caption{Results of our method with the three different protocols.}
    \begin{adjustbox}{width=0.489\textwidth}
    \begin{tabular}{l|c|c|c}
    \textbf{Protocol} & \textbf{\% of frames} & \textbf{MAE} & \textbf{RMSE} \\ \hline \hline
    \multirow{2}{*}{Leave-One-Sequence-Out} & 25\% & \textbf{2.3166} & 3.1459 \\ \cline{2-4}
     & 100\% & 2.5291 & 3.3263 \\ \Xhline{2\arrayrulewidth}
    \multirow{2}{*}{Leave-One-Subject-Out cross validation} & 25\% & \textbf{2.523} & 3.2692 \\ \cline{2-4}
     & 100\% & 2.9176 & 3.5133 \\ \Xhline{2\arrayrulewidth}
    \multirow{2}{*}{5-fold cross validation} & 25\% & \textbf{2.4365} & 3.147 \\ \cline{2-4}
     & 100\% & 2.7944 & 3.5088 \\ \Xhline{2\arrayrulewidth}
    \end{tabular}
    \end{adjustbox}
    \label{tab:results-velocities}
\end{table}

\begin{figure}[!ht]
\centering
\includegraphics[width=0.8\linewidth]{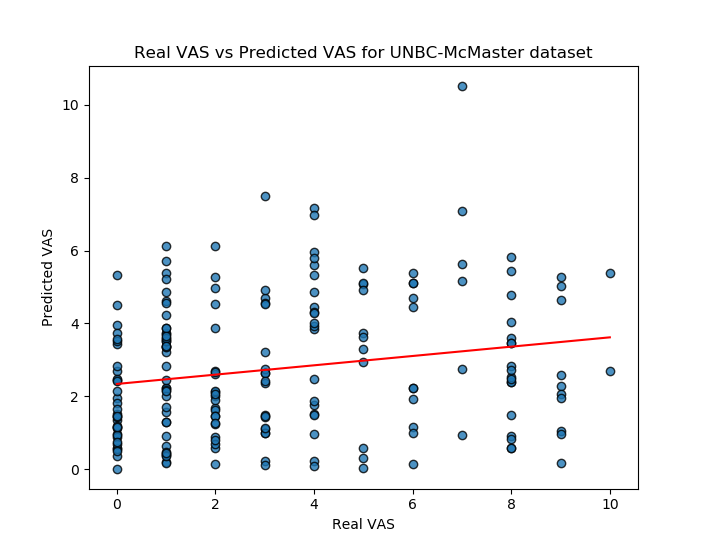}
\caption{Distribution of the predicted VAS values compared with the real VAS using the 5-fold cross validation protocol with 25\% of frames. The red line is a least-square fitting of the predicted values.}
\label{fig:distribution-pain-estimated}
\end{figure}

From Table~\ref{tab:results-velocities}, we notice that in every cases, the MAE is lower when we down-sample 1 frame each 4 frames leading to 25\% of the frames available for pain assessment. 
This is due to the high amount of non-pain frames that are present in the dataset. We also notice that the best MAE we obtained is 2.3166 with the Leave-One-Sequence-Out protocol. This result is expected as this protocol is not subject-independent and sequences of the same subject can be used for both training and testing. The second best MAE we obtained is 2.4365, using the 5-fold cross validation protocol. We report the RMSE as a second measure of the error of our estimation. Results show the same trend as the MAE with the best RMSE observed for the Leave-One-Sequence-Out protocol. 

Similar to~\cite{LiuPRP17-DeepFaceLIFT}, we present in Fig.~\ref{fig:distribution-pain-estimated} the distribution of the predicted VAS score against the true reported VAS. This allows us to observe that our approach is capable of predicting many low VAS scores and can have difficulty in estimating higher values.

\subsection{Comparison with state-of-the-art}
We compared our approach to the two state-of-the-art methods for VAS pain intensity measurement from video (see Table~\ref{tab:comparison-results-sota}). Here, we report the best results for DeepFaceLIFT~\cite{LiuPRP17-DeepFaceLIFT} that only uses the VAS as training labels as the authors also present results while combining VAS and OPR labels. They obtained a MAE of 2.30 using a 5-fold cross validation protocol. Our results are close to theirs, while only using a geometry based formulation of facial landmark dynamics (meaning that our method is less expensive as we do not have to train a neural network). Our results are comparable to RNN-HCRF~\cite{Martinez:cvrpw17} results, as they obtain a MAE of 2.46, though using a different protocol. In fact, in the results for RNN-HCRF, data have been randomly split by taking the sequences of 15 subjects for training and the sequences of 10 subjects for testing. It is also important to highlight that in RNN-HCRF the face appearance is also used, while our method only considers the shape of the face.

\begin{table}[!htb]
\centering
\caption{Comparison of our method with state-of-the-art results}
    \begin{adjustbox}{width=0.489\textwidth}
    \begin{tabular}{l|c|c|c}
        \textbf{Method} & \textbf{Protocol} & \textbf{Labels for training} & \textbf{MAE} \\ \hline \hline
        DeepFaceLift~\cite{LiuPRP17-DeepFaceLIFT} & 5-fold cross validation & VAS & 2.30 \\ \hline
        RNN-HCRF~\cite{Martinez:cvrpw17} & random split & VAS \& PSPI & 2.46 \\ \hline \hline
        Ours & 5-fold cross validation & VAS & 2.4365 \\ \hline
    \end{tabular}
    \end{adjustbox}
    \label{tab:comparison-results-sota} 
\end{table}


\section{Conclusion}\label{sec:conclusions}
We proposed a method based on facial landmarks dynamics to estimate pain intensity from video. Our approach shows competitive results with respect to state-of-the-art methods on the UNBC-McMaster Shoulder Pain Archive, while only considering the shape of the face. 
Future work will focus on the combination of facial shape and appearance as well as the inclusion of other pain scales such as the observer pain rating scale to further improve the pain scores estimation.

\section*{Acknowledgment}
Zakia Hammal's effort was supported by the National Institute of Nursing Research of the National Institutes of Health under Awards Number R21NR016510 and R01NR018451. The content is solely the responsibility of the authors and does not necessarily represent the official views of the National Institutes of Health. The proposed work was also partially supported by the French State, managed by the National Agency for Research (ANR) under the Investments for the future program with reference ANR-16-IDEX-0004 ULNE. We thank Prof. J-C. Alvarez Paiva from University of Lille for fruitful discussions on the formulation of the distance between $n \times 2$ landmark configurations in Eq.~\eqref{eq:distp2}. We also thank Dr. Pierre-Yves Gousenbourger from Université Catholique de Louvain for providing us the curve fitting Matlab code.



%





\begin{thebibliography}{10}
\providecommand{\url}[1]{#1}
\csname url@samestyle\endcsname
\providecommand{\newblock}{\relax}
\providecommand{\bibinfo}[2]{#2}
\providecommand{\BIBentrySTDinterwordspacing}{\spaceskip=0pt\relax}
\providecommand{\BIBentryALTinterwordstretchfactor}{4}
\providecommand{\BIBentryALTinterwordspacing}{\spaceskip=\fontdimen2\font plus
\BIBentryALTinterwordstretchfactor\fontdimen3\font minus
  \fontdimen4\font\relax}
\providecommand{\BIBforeignlanguage}[2]{{%
\expandafter\ifx\csname l@#1\endcsname\relax
\typeout{** WARNING: IEEEtran.bst: No hyphenation pattern has been}%
\typeout{** loaded for the language `#1'. Using the pattern for}%
\typeout{** the default language instead.}%
\else
\language=\csname l@#1\endcsname
\fi
#2}}
\providecommand{\BIBdecl}{\relax}
\BIBdecl

\bibitem{Merskey:1979}
H.~Merskey and et~al., ``Pain terms: a list with definitions and notes on
  usage,'' \emph{Pain}, vol.~6, no.~3, 1979.

\bibitem{Younger2009PainOA}
J.~Younger, R.~McCue, and S.~Mackey, ``Pain outcomes: A brief review of
  instruments and techniques,'' \emph{Current Pain and Headache Reports},
  vol.~13, pp. 39--43, 2009.

\bibitem{Aicher:2012}
B.~Aicher, H.~Peil, B.~Peil, and H.-C. Diener, ``Pain measurement: Visual
  analogue scale (vas) and verbal rating scale (vrs) in clinical trials with
  otc analgesics in headache,'' \emph{Cephalalgia}, vol.~32, no.~3, pp.
  185--197, 2012.

\bibitem{Farrar:2000}
J.~T. Farrar, R.~K. Portenoy, J.~A. Berlin, J.~L. Kinman, and B.~L. Strom,
  ``Defining the clinically important difference in pain outcome measures,''
  \emph{Pain}, vol.~88, no.~3, pp. 287--294, 2000.

\bibitem{Jensen:2003}
M.~P. Jensen, C.~Chen, and A.~M. Brugger, ``Interpretation of visual analog
  scale ratings and change scores: a reanalysis of two clinical trials of
  postoperative pain,'' \emph{The Journal of Pain}, vol.~4, no.~7, pp.
  407--414, 2003.

\bibitem{Jensen:2005}
M.~P. Jensen, S.~A. Martin, and R.~Cheung, ``The meaning of pain relief in a
  clinical trial,'' \emph{The Journal of Pain}, vol.~6, no.~6, pp. 400--406,
  2005.

\bibitem{Craig:2011}
K.~D.~C. et~al., \emph{The facial expression of pain}.\hskip 1em plus 0.5em
  minus 0.4em\relax Guilford Press, 2011.

\bibitem{Kunz:2007}
M.~Kunz, S.~Scharmann, U.~Hemmeter, K.~Schepelmann, and S.~Lautenbacher, ``The
  facial expression of pain in patients with dementia,'' \emph{Pain}, vol. 133,
  no.~1, 2007.

\bibitem{Prkachin:1992}
K.~M. Prkachin, ``The consistency of facial expressions of pain: a comparison
  across modalities,'' \emph{Pain}, vol.~51, no.~3, 1992.

\bibitem{Prkachin:2008}
K.~M. Prkachin and P.~E. Solomon, ``The structure, reliability and validity of
  pain expression: Evidence from patients with shoulder pain,'' \emph{Pain},
  vol. 139, no.~2, pp. 267--274, 2008.

\bibitem{Ekman:2002}
P.~Ekman, W.~Friesen, and J.~Hager, \emph{Facial Action Coding System: The
  Manual on CD ROM}, 2002.

\bibitem{Hammaletal2018}
Z.~Hammal and J.~F. Cohn, ``Automatic, objective, and efficient measurement of
  pain using automated face analysis,'' \emph{Ken Prkachin, Zina Trost and Kai
  Karos (EDs.), Handbook of Social and interpersonal processes in pain: We
  don’t suffer alone}, p. 121–146, 2018.

\bibitem{Werner:2019}
P.~{Werner}, D.~{Lopez-Martinez}, S.~{Walter}, A.~{Al-Hamadi}, S.~{Gruss}, and
  R.~{Picard}, ``Automatic recognition methods supporting pain assessment: A
  survey,'' \emph{IEEE Trans. on Affective Computing}, pp. 1--1, to appear
  2019.

\bibitem{Martinez:cvrpw17}
D.~L. Martinez, O.~Rudovic, and R.~W. Picard, ``Personalized automatic
  estimation of self-reported pain intensity from facial expressions,'' in
  \emph{{IEEE} Conf. on Computer Vision and Pattern Recognition Workshops
  {CVPR}}, 2017, pp. 2318--2327.

\bibitem{LiuPRP17-DeepFaceLIFT}
D.~Liu, F.~Peng, O.~O. Rudovic, and R.~W. Picard, ``Deepfacelift: Interpretable
  personalized models for automatic estimation of self-reported pain,'' in
  \emph{AffComp@IJCAI}, ser. Proceedings of Machine Learning Research,
  vol.~66.\hskip 1em plus 0.5em minus 0.4em\relax {PMLR}, 2017, pp. 1--16.

\bibitem{UNBC-McMaster}
P.~Lucey, J.~F. Cohn, K.~M. Prkachin, P.~E. Solomon, and I.~A. Matthews,
  ``Painful data: The unbc-mcmaster shoulder pain expression archive
  database,'' in \emph{{IEEE} Int. Conf. on Automatic Face and Gesture
  Recognition {(FG)}}, 2011, pp. 57--64.

\bibitem{Szczapa2019ICCVW}
B.~Szczapa, M.~Daoudi, S.~Berretti, A.~Del~Bimbo, P.~Pala, and E.~Massart,
  ``Fitting, comparison, and alignment of trajectories on positive
  semi-definite matrices with application to action recognition,'' in
  \emph{IEEE Int. Conf. on Computer Vision (ICCV) Workshops}, Oct 2019.

\bibitem{DruckerBKSV96SVR}
H.~Drucker, C.~J.~C. Burges, L.~Kaufman, A.~J. Smola, and V.~Vapnik, ``Support
  vector regression machines,'' in \emph{Advances in Neural Information
  Processing Systems (NIPS)}, 1996, pp. 155--161.

\bibitem{KacemDABP20}
\BIBentryALTinterwordspacing
A.~Kacem, M.~Daoudi, B.~{Ben Amor}, S.~Berretti, and J.~C. {Alvarez Paiva}, ``A
  novel geometric framework on gram matrix trajectories for human behavior
  understanding,'' \emph{{IEEE} Trans. Pattern Analysis and Machine
  Intelligence}, vol.~42, no.~1, pp. 1--14, 2020. [Online]. Available:
  \url{https://doi.org/10.1109/TPAMI.2018.2872564}
\BIBentrySTDinterwordspacing

\bibitem{bonnabel:2009}
S.~Bonnabel and R.~Sepulchre, ``Riemannian metric and geometric mean for
  positive semidefinite matrices of fixed rank,'' \emph{SIAM J. Matrix Anal.
  Appl.}, vol.~31, no.~3, pp. 1055--1070, 2009.

\bibitem{journee2010low}
M.~Journ{\'e}e, F.~Bach, P.-A. Absil, and R.~Sepulchre, ``Low-rank optimization
  on the cone of positive semidefinite matrices,'' \emph{SIAM Journal on
  Optimization}, vol.~20, no.~5, pp. 2327--2351, 2010.

\bibitem{massart2020quotient}
E.~Massart and P.-A. Absil, ``Quotient geometry with simple geodesics for the
  manifold of fixed-rank positive-semidefinite matrices,'' \emph{SIAM Journal
  on Matrix Analysis and Applications}, vol.~41, no.~1, pp. 171--198, 2020.

\bibitem{massart2019curvature}
E.~Massart, J.~M. Hendrickx, and P.-A. Absil, ``Curvature of the manifold of
  fixed-rank positive-semidefinite matrices endowed with the
  {Bures-Wasserstein} metric,'' in \emph{4th Conference on Geometric Sciences
  of Information (GSI 2019)}, 2019, pp. 739--748.

\bibitem{vandereycken2009embedded}
B.~Vandereycken, P.-A. Absil, and S.~Vandewalle, ``Embedded geometry of the set
  of symmetric positive semidefinite matrices of fixed rank,'' in \emph{IEEE/SP
  Workshop on Statistical Signal Processing (SSP)}, 2009, pp. 389--392.

\bibitem{vandereycken2013riemannian}
B.~{Vandereycken}, P.-A. {Absil}, and S.~{Vandewalle}, ``A {R}iemannian
  geometry with complete geodesics for the set of positive semidefinite
  matrices of fixed rank,'' \emph{IMA Journal of Numerical Analysis}, vol.~33,
  no.~2, pp. 481--514, 2013.

\bibitem{faraki:2016}
M.~Faraki, M.~T. Harandi, and F.~Porikli, ``Image set classification by
  symmetric positive semi-definite matrices,'' in \emph{IEEE Winter Conf. on
  Applications of Computer Vision (WACV)}, 2016, pp. 1--8.

\bibitem{meyer2011regression}
G.~Meyer, S.~Bonnabel, and R.~Sepulchre, ``Regression on fixed-rank positive
  semidefinite matrices: a {R}iemannian approach,'' \emph{Journal of Machine
  Learning Research}, vol.~12, no. Feb, pp. 593--625, 2011.

\bibitem{gousenbourger2017piecewise}
P.-Y. Gousenbourger, E.~Massart, A.~Musolas, P.-A. Absil, L.~Jacques, J.~M.
  Hendrickx, and Y.~Marzouk, ``Piecewise-{B\'e}zier {$C^1$} smoothing on
  manifolds with application to wind field estimation,'' 2017, pp. 305--310.

\bibitem{massart2019interpolation}
E.~Massart, P.-Y. Gousenbourger, N.~T. Son, T.~Stykel, and P.-A. Absil,
  ``{Interpolation on the manifold of fixed-rank positive-semidefinite matrices
  for parametric model order reduction: preliminary results},'' in
  \emph{European Symposium on Artifical Neural Networks, Computational
  Intelligence and Machine Learning (ESANN)}, 2019, pp. 281--286.

\bibitem{Gousenbourger2018}
P.-Y. Gousenbourger, E.~Massart, and P.-A. Absil, ``Data fitting on manifolds
  with composite {B\'e}zier-like curves and blended cubic splines,''
  \emph{Journal of Mathematical Imaging and Vision}, vol.~61, no.~5, pp.
  645--671, 2018.

\bibitem{CuturiVBM07}
M.~Cuturi, J.~Vert, {\O}.~Birkenes, and T.~Matsui, ``A kernel for time series
  based on global alignments,'' in \emph{{IEEE} Int. Conf. on Acoustics,
  Speech, and Signal Processing {ICASSP}}, 2007, pp. 413--416.

\bibitem{manopt}
\BIBentryALTinterwordspacing
N.~Boumal, B.~Mishra, P.-A. Absil, and R.~Sepulchre, ``Manopt, a matlab toolbox
  for optimization on manifolds,'' \emph{Journal of Machine Learning Research},
  vol.~15, pp. 1455--1459, 2014. [Online]. Available:
  \url{http://jmlr.org/papers/v15/boumal14a.html}
\BIBentrySTDinterwordspacing

\end{thebibliography}


\end{document}